\documentclass[a4paper]{article}
\usepackage[a4paper,left=2.5cm,right=2.5cm,top=2.5cm,bottom=2.5cm]{geometry}
\usepackage{amsmath,amsfonts,bm}
\usepackage{algorithmic}
\usepackage{algorithm}
\usepackage{authblk}
\usepackage{array}
\usepackage[caption=false,font=normalsize,labelfont=sf,textfont=sf]{subfig}
\usepackage{textcomp}
\usepackage{stfloats}
\usepackage{url}
\usepackage{verbatim}
\usepackage{graphicx}
\usepackage{cite}
\usepackage{nicefrac}       
\usepackage{microtype}      
\usepackage{xcolor}         
\usepackage{multirow}
\usepackage{booktabs}       

\title{P$^2$U: Progressive Precision Update For Efficient Model Distribution}

\author[a]{Homayun Afrabandpey}
\author[b]{Hamed Rezazadegan Tavakoli}

\affil[a]{homayun.it@gmail.com}
\affil[b]{hamed.rezazadegan\_tavakoli@nokia.com}

\date{}

\begin{document}

\maketitle

\begin{abstract}
Efficient model distribution is becoming increasingly critical in bandwidth-constrained environments. In this paper, we propose a simple yet effective approach called Progressive Precision Update (P\textsuperscript{2}U) to address this problem. Instead of transmitting the original high-precision model, P\textsuperscript{2}U transmits a lower-bit precision model, coupled with a model update representing the difference between the original high-precision model and the transmitted low precision version. With extensive experiments on various model architectures, ranging from small models ($1 - 6$ million parameters) to a large model (more than $100$ million parameters) and using three different data sets, e.g., chest X-Ray, PASCAL-VOC, and CIFAR-100, we demonstrate that P\textsuperscript{2}U consistently achieves better tradeoff between accuracy, bandwidth usage and latency. Moreover, we show that when bandwidth or startup time is the priority, aggressive quantization (e.g., 4-bit) can be used without severely compromising performance. These results establish P\textsuperscript{2}U as an effective and practical solution for scalable and efficient model distribution in low-resource settings, including federated learning, edge computing, and IoT deployments. Given that P\textsuperscript{2}U complements existing compression techniques and can be implemented alongside any compression method, e.g., sparsification, quantization, pruning, etc., the potential for improvement is even greater.
\end{abstract}


\section{Introduction}\label{sec:intro}
With the increasing complexity and explosive growth in the size of Machine Learning (ML) models, efficient model transmission plays a key role in applications with distributed and resource-constrained nature such as Internet of Things (IoT) and Federated Learning (FL). Transmitting large models in their original high-bit precision, e.g., 32-bit float, across networks with limited bandwidth can result in substantial communication overhead while maintaining the highest possible accuracy. Another issue is the increased startup latency, i.e., the time it takes for the receiver to start inference. Increasing startup latency can have detrimental effects on a user's Quality of Experience (QoE). As an example, assume a user starts an app on their edge device, e.g., mobile phone, to perform real-time object detection in the video stream captured by the device camera, for example in an Augmented Reality (AR) use case. The app may use a pre-trained model for the task. An example of such a model may be VGG16 \cite{simonyan2015very}. A pre-trained VGG16 can be up to $528$ Mega Bytes (MB) in size, which can take substantial time to be downloaded from a service provider model repository, even if the repository is hosted in the 5G network. For example, on a link with an average capacity of $100$Mbps with even a low access latency of $10$ms, it may take up to a minute for the model to be downloaded. One minute of startup latency can be quite detrimental in a latency-sensitive scenario such as AR.

In response to this issue, a large body of literature exists on different types of compression techniques including sparsification \cite{fedorov2019sparse,sattler2019sparse,wangni2018gradient}, pruning \cite{liu2021compressing,qi2021efficient}, quantization \cite{goutham2022stochastic,li2020fixed,yang2021communication} and encoding \cite{Deep2016Han,wiedemann2020deepcabac}. These techniques generally apply transformations to the model weights or architecture to reduce the model size before transmission, aiming to achieve a smaller model that can be transmitted in one go. This way they sacrifice accuracy for bandwidth usage while also decreasing startup latency. In this paper, looking at the problem from a different angle, we propose a novel approach called Progressive Precision Update (P\textsuperscript{2}U) that focuses on model compression during the transmission process, not prior to that. P\textsuperscript{2}U centers on the concept of precision-driven model transmission. Instead of sending the full precision model (compressed or uncompressed) directly, we advocate for the initial transmission of a lower-bit precision version of the model, for instance, an 8-bit integer model. This lower-precision model serves as a compressed representation that captures the essential characteristics of the original model while drastically reducing the amount of transmitted data. The receiver can start inference upon receiving this low-precision model. To bridge the accuracy degradation due to precision gap, we then transmit and apply model update(s), representing the discrepancy between the original high-precision model and its lower-precision counterpart. The motivation behind this approach is three-fold. First, it addresses the bandwidth limitations inherent in distributed learning scenarios, providing a more lightweight alternative for model transmission. Second, the reduction in the model size and effective information transfer leads to decreased startup latency time. This makes our approach particularly suitable for real-time applications. Third, P\textsuperscript{2}U provides fine-grained control over the transmission process, allowing for tailored updates based on available bandwidth or specific accuracy requirements. It is important to note that P\textsuperscript{2}U is complementary to model compression techniques and could be used in conjunction with them to further improve bandwidth saving.

Our method is particularly effective at navigating the tradeoff between accuracy, bandwidth usage, and startup latency. For instance, in the case of MobileNet-v2 and VGG16 on the PASCAL-VOC dataset, P\textsuperscript{2}U with an 8-bit low-precision model outperforms even the 16-bit quantized baseline in terms of accuracy, while maintaining significantly lower communication cost and startup time. This confirms that the transmitted lower-precision model acts as an efficient approximation, capturing essential features of the original model. We present a comprehensive analysis of our method using diverse data sets and model architecture settings, showcasing its efficacy across various scenarios. Our findings emphasize the potential of precision-driven model transmission as a viable solution for bandwidth optimization in distributed learning, paving the way for more scalable and responsive systems in resource-constrained environments.

\section{Related Works}\label{sec:related_works}
Several lines of research have addressed the challenge of efficient model distribution and inference in bandwidth-constrained or latency-sensitive environments. Broadly, existing approaches can be categorized into model compression techniques, progressive inference schemes, and update-efficient model distribution methods. However, none of these methods explicitly target all three axes of efficiency -- accuracy, bandwidth, and startup time -- as jointly and effectively as our proposed P\textsuperscript{2}U method.

Model quantization and compression techniques reduce bandwidth and memory requirements by lowering the precision of weights and activations. The quantization~\cite{nagel2021white} may happen in various forms ranging from the known signal processing approaches, e.g., uniform quantization, and alike; to more advanced techniques such as parameter and data scaling~ \cite{michaud2023the}. In scaling approaches, a weight matrix may be decomposed to a scale vector and a quantized matrix, allowing more accurate recovery or better performance at the inference time. The scaling approaches are more common when a parameter sensitive component is processed, e.g., batchnorm folding. While these methods achieve good compression ratios, they offer limited flexibility: once the model is quantized to a certain precision level, it must be distributed and used as-is. They do not support progressive quality upgrades or adaptive tradeoffs between accuracy and transmission cost at runtime. A P\textsuperscript{2}U related line of research in this area is quantization with adaptive bit-widths \cite{jin2020adabits}. This approach aims at the adaptive execution of models at different bit-widths and requires training quantized neural networks with different precisions adaptive to different requirements, while our method aims for efficient model distribution and inference.

Progressive transmission methods, such as BitSplit~\cite{wang2020towards} and progressive pruning~\cite{ye2018progressive}, aim to deliver model components in multiple stages to gradually refine performance. While they support staged deployment, they often require specialized architectures or substantial retraining. Moreover, they rarely account for the importance of minimizing startup latency, as they focus on full model reconstruction eventually. In contrast, P\textsuperscript{2}U is model-agnostic and designed to enable useful inference even from a low-precision base model, allowing quick deployment followed by progressive refinement.

Pruning and sparsification are two other extensively researched techniques in recent years \cite{NIPS2016_Lee, singh2020woodfisher, tanaka2020pruning, tavakoli2021hybrid, tung2018deep, yeom2021pruning}. Matrix decomposition based approaches are yet another favored approach for reducing the size of a neural network for transfer. Familiar with matrix decomposition methods from linear algebra, one may decompose a weight tensor into smaller matrices. Various works have explored this aspect \cite{chen2023joint, liu2024deep, swaminathan2020sparse}.

The demand for neural model compression is significant that there has been standardization activities related to the topic of neural compression~\cite{9478787}. The neural model compression standard ensures an interoperable mechanism for delivering compressed neural networks. The pipeline consists of means for sparsification, pruning, quantization, and entropy coding of a model. It furthermore ensures mechanisms for encoding incremental updates, similar to federated learning, are supported. The proposed approach builds on top of the capabilities of neural model compression standard to demonstrate effectiveness of incremental delivery of a compressed model.

Another related family of methods includes model distillation and proxy-based distribution approaches~\cite{polino2018model,sanh2019distilbert}, which use smaller student models or surrogate networks to reduce deployment cost. These approaches typically prioritize inference efficiency but often sacrifice accuracy or require expensive teacher-student training pipelines. P\textsuperscript{2}U, by contrast, uses a single model architecture and progressively improves it without retraining.

A relevant, yet distinct, domain is compression of federated communications~\cite{mcmahan2023communicationefficient}. These methods explore how to transmit model updates efficiently under communication constraints. For example,~\cite{JinHyun2019, Seungeun2020} consider communication channel characteristics during federated learning to reduce the amount of bits required for transmitting a model. Other approaches to compressing federated communications may follow the principles of compression of neural networks for deployment, i.e., sparsification/pruning, quantization, and entropy coding, e.g.,~\cite{afrabandpey2022importance, Becking2024, Yongjeong2021}. Our approach may share conceptual commonalities with these methods where weight updates are communicated between entities. While they optimize bandwidth during training across distributed clients, they do not directly address startup latency or allow inference from intermediate representations. P\textsuperscript{2}U brings these ideas to the inference stage by allowing the user to begin working with a low-precision model almost immediately, followed by incremental improvements as higher-precision updates are received.

Another related concept is incremental model transfer, which is defined as determining a split point that allows partial transfer of the model and partial execution of the computational graph akin to the split learning and inference~\cite{GUPTA20181} until the complete model is transferred and executed. P\textsuperscript{2}U is orthogonal to the incremental model transfer. In our approach, the model is deployed completely at different bit increments that allow enabling bit-incremental model deployment.

In summary, while prior works make important contributions toward reducing model size, bandwidth usage, or inference cost, they typically optimize along a single dimension. In contrast, P\textsuperscript{2}U is designed to operate in a tri-objective space: achieving a balance between accuracy, bandwidth efficiency, and startup latency. As demonstrated in our experiments, P\textsuperscript{2}U enables flexible deployment strategies, supporting rapid startup with coarse models and seamless refinement toward high-accuracy solutions -- making it particularly suitable for resource -- constrained or latency-sensitive applications.

\section{Method}\label{sec:method}
Imagine a smart city equipped with an advanced traffic management system that leverages ML to optimize traffic flow, reduce congestion, and minimize environmental impact \cite{lv2014traffic,ma2015large}. As a new car enters the city, equipped with IoT sensors and communication capabilities, it initiates communication with the central traffic management server. The car's onboard systems seek a ML model that can dynamically adapt to real-time traffic conditions, predict congestion patterns, and optimize the vehicle's navigation route to reduce travel time and energy consumption. In this scenario, a big challenge lies in the efficient transmission of the ML model to the new vehicle, considering potential bandwidth limitations and the need for rapid inference. We propose P\textsuperscript{2}U with the motivation of bandwidth saving during model transmission and deployment and decreasing startup latency in latency critical scenarios. 

Since compressed model transmission is usually done in the quantized domain, we assume that the model requested by the edge device is either available in the server in quantized form with different precisions or the server is able to quantize the model to the required precisions. Figure \ref{fig:p2u_workflow} graphically demonstrates the workflow of P\textsuperscript{2}U.
P\textsuperscript{2}U works in steps as follows:

\begin{figure*}[t!]
    \centering
    \includegraphics[scale=0.35]{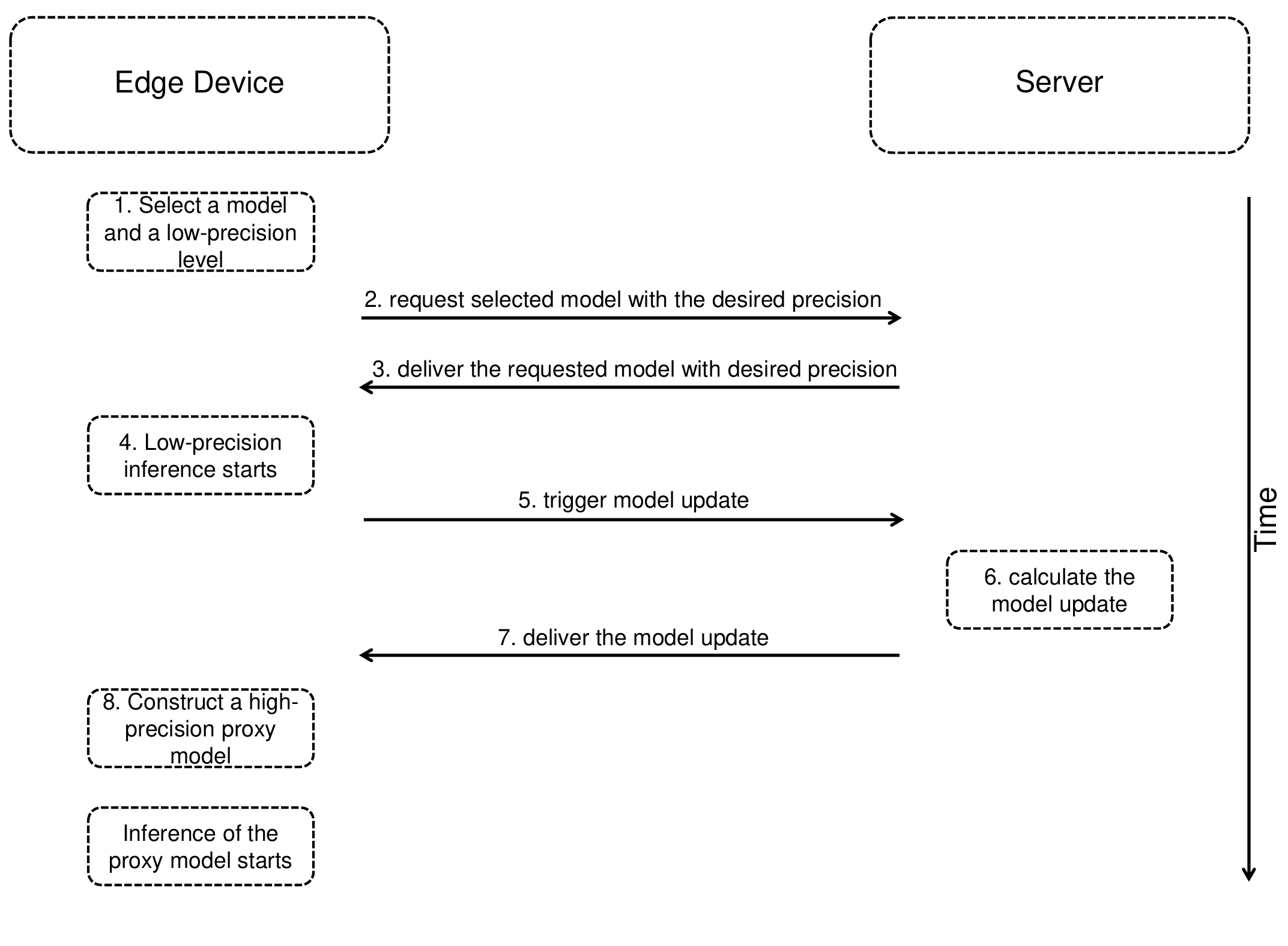}
    \caption{Workflow of P\textsuperscript{2}U. After receiving the request for a ML model with a specific low-precision level from the edge device, the server first delivers the model with the requested precision level and then sends an update as the difference between the original (high-precision) model and the delivered low-precision model.}
    \label{fig:p2u_workflow}
\end{figure*}

\begin{itemize}
    \item[1.] a model and a desired low-precision level is selected by the edge device based on, e.g., device resources, network conditions, etc.,
    \item[2.] the edge device sends a request for the selected model with the desired low-precision level to the server,
    \item[3.] the server identifies the selected model in its model repository and prepares the requested precision if it is not already available in the repository and sends the model with the requested precision to the edge device, 
    \item[4.] the edge device starts inferencing upon receiving the low-precision model,
    \item[5.] the edge device triggers a model update (parallel to the inferencing process of step 4). The update is a precision update of the model currently at the edge device rather than a new model,
    \item[6.] the server computes the model update as the difference between the original high-precision and transferred low-precision model,
    \item[7.] the model update is sent to the edge device,
    \item[8.] the edge device constructs a proxy to the original high-precision model in the server by adding the update to the low-precision model and adopts this high-precision proxy for the inference.
\end{itemize}
In step 5, the edge device may trigger a model update immediately after receiving the model or it may trigger the update after a period of time. The update may be triggered based on a variety of factors including, but not limited to, model update delivery time, accuracy achieved at the edge device using the low-precision model, prospective accuracy improvement achievable with a model update, change in accuracy requirements, etc. The following subsection provides a theoretical proof that, under certain assumptions, the accuracy loss between the original high-precision model and its proxy constructed in the edge device by adding the update to the low-precision model, is minimal.

\subsection{Theoretical Analysis}\label{subsec:theory}
Consider a neural network with high precision, e.g., 32-bit int, and low precision, e.g., 8-bit int, quantized weights, denoted as $\bm{W}^h_Q$ and $\bm{W}^l_Q$, respectively. Let $\bm{W}^h$ and $\bm{W}^l$ be the corresponding dequantized 32-bit float versions of these weights. The update is defined as:
\begin{equation}\label{eq:update}
    \Delta\bm{W}=\bm{W}^h - \bm{W}^l,
\end{equation}
with its quantized version being shown as $\Delta\bm{W}_Q$. At the receiver, we construct a high-precision proxy for $\bm{W}^h$ as $\bm{W}'=\bm{W}^l+\Delta\bm{W}$. Given an input $\bm{x}$, the outputs of the high-precision model and low-precision model are denoted as $f\left(\bm{W}^h,\bm{x}\right)$ and $f\left(\bm{W}^l,\bm{x}\right)$, respectively. The error induced by using the low-precision model for inference is:
\begin{equation}\label{eq:error}
    \epsilon\left(x\right) = f\left(\bm{W}^h,\bm{x}\right) - f\left(\bm{W}^l,\bm{x}\right).
\end{equation}
Our goal is to show that the high-precision proxy model $\bm{W}'$ compensates for the error $\epsilon\left(\bm{x}\right)$. In other words, we aim to show that $f\left(\bm{W}',\bm{x}\right)\approx f\left(\bm{W}^h,\bm{x}\right)$. 

We analyse the behavior of the function $f$ using a first-order Taylor expansion. Assuming $f$ is smooth and differentiable, we expand $f\left(\bm{W}^h,\bm{x}\right)$ around $\bm{W}^l$:
\begin{equation}\label{eq:taylor_orig}
    f\left(\bm{W}^h, \bm{x}\right)\approx f\left(\bm{W}^l,\bm{x}\right) + \nabla f\left(\bm{W}^l,\bm{x}\right).\left(\bm{W}^h-\bm{W}^l\right) + R_h,
\end{equation}
where $R_h$ represents high-order residual terms. Similarly, expanding $f\left(\bm{W}', \bm{x}\right)$ around $\bm{W}^l$, we get:
\begin{equation}\label{eq:taylor_recon}
    \begin{aligned}
        f\left(\bm{W}',\bm{x}\right) &= f\left(\bm{W}^l+\Delta\bm{W},\bm{x}\right) \approx f\left(\bm{W}^l,\bm{x}\right) + \\
        &\nabla f\left(\bm{W}^l,\bm{x}\right) \cdot \left(\bm{W}^h-\bm{W}^l\right) + R'.
    \end{aligned}
\end{equation}
Given \ref{eq:taylor_orig} and \ref{eq:taylor_recon}, we obtain:
\begin{equation*}
    f\left(\bm{W}',\bm{x}\right)\approx f\left(\bm{W}^h,\bm{x}\right) + \left(R'-R_h\right).
\end{equation*}
Considering these, the difference between the output of the proxy model and the high-precision model obtains as:
\begin{equation}
	|f\left(\bm{W}',\bm{x}\right)-f\left(\bm{W}^h,\bm{x}\right)|\leq|R'-R_h|.
\end{equation}
Assuming\footnote{the first assumption ensures the validity of first-order Taylor approximation, and the second assumption ensures that higher-order terms diminish quadratically, making their effect negligible for small $\delta$}:
\begin{itemize}
	\item[1.] $\parallel \bm{W}^h-\bm{W}^l\parallel\leq\delta$ where $\delta$ is relatively small, and
	\item[2.] $\parallel R_h\parallel, \parallel R'\parallel\leq \frac{M}{2}\parallel \bm{W}^h-\bm{W}^l\parallel^2$,
\end{itemize}
since $|R'-R_h|$ is on the order of $O\left(\delta^2\right)$, the proxy model approximates the high-precision model with high fidelity, leading to negligible inference error in practice.

\section{Experiments}\label{sec:expriments}
\subsection{Experimental Setup}
In this subsection, we provide detailed settings of our experiments necessary for reproducing the results.
\subsubsection{Data sets}\label{subsec:dataset}
We perform a set of experiments on three standard image classification data sets:
\begin{itemize}
    \item Chest X-Ray \cite{kermany2018large} is a binary image classification data set with in total $5856$ images from which $5232$ images are used for training and the remaining $624$ images are used as test data. From all images in this data set, $1583$ images are ``Normal'' and the rest are ``Pneumonia'' cases. 
    \item PASCAL-VOC \cite{pascal-voc-2012} version 2012, which contains in total $11540$ images from $20$ different classes. We adopted a pre-defined split of the data set with $6925$ images for training, $2308$ images for testing, and the remaining $2307$ images for validation.\footnote{although, we have not used validation data in our implementation.}
    \item CIFAR-100 \cite{krizhevsky2009learning} consists of $60000$ color images of size $32\times32$ from $100$ different classes. There are $50000$ training images, i.e., $500$ training images per class, and $10000$ test images, i.e., $100$ test images per class.
\end{itemize}

\subsubsection{Network Architectures}\label{subsec:net_arch}

We used MobileNet-v2, ResNet18, EfficientNet-b4, and VGG16, all pre-trained on ImageNet and loaded from PyTorch model zoo. Table \ref{Tab:architecture_info} demonstrates the number of parameters (in millions) and the model sizes (in MB) when loaded from the PyTorch model zoo. Since these models were originally trained on ImageNet with $224\times224$ images and $1000$ classes, directly applying them to data sets with differing number of classes leads to reduced accuracy. To address this, we modified each model by adjusting the output dimension of the final fully connected layer to match the number of classes in the target dataset and when necessary (in the case of CIFAR-100 where image sizes are much smaller than ImageNet) also modified initial feature layer. Finally, to enhance performance, we retrained each model on the training set of the selected dataset before initiating model transfer. During this process, all parameters except those modified in the transformation step were frozen.

\begin{table}[]
    \caption{Number of parameters (in Million) and Size (MB) of each model architecture used in the experiment.}
    \centering
    \begin{tabular}{|l|c|c|c|c|}
        \hline
        & \textbf{MobileNet-v2} & \textbf{ResNet18} & \textbf{EfficientNet-b4} & \textbf{VGG16} \\ \hline
        \textbf{Parameters} & 3.5 & 11.7 & 19.4 & 138.4 \\ \hline
        \textbf{Size} & 14 & 41 & 74.5 & 489 \\ \hline
    \end{tabular}
    \label{Tab:architecture_info}
\end{table}

\subsubsection{Implementation}\label{subsec:implementation}
We used PyTorch framework for the implementation. As a preprocessing step for the Chest X-Ray data set, we resized the training images into $150\times150$ and applied a random affine transformation\footnote{without rotation (``degree'' = 0) with scale in the range $\left(0.8, 1.2\right)$}. Test images are also resized to $150\times150$ without any other transformation. In the preprocessing of the PASCAL-VOC data set, when using MobileNet-v2, ResNet18, and VGG16, both train and test images are resized into $256\times 256$, center cropped by $224\times 224$, and normalized to have zero mean and a standard deviation of $1$. In the special case where EfficientNet-b4 is used with PASCAL-VOC, train and test images are resized into $412\times 412$, center cropped by $380\times 380$, and normalized to have zero mean and a standard deviation of $1$. This is because naturally, EfficientNet expects larger images and might not work well with datasets with small images without resizing to larger sizes or making other adjustments to the architecture. Finally, the preprocessing step of CIFAR-100 includes normalization of both training and test data to zero mean and a standard deviation of $1$. For quantization and entropy coding, we used NNCodec software \cite{becking2023nncodec} that implements the standard-compliant implementation of the Neural Network Coding (NNC) standard (ISO/IEC 15938-17) \cite{kirchhoffer2021overview}.

For model training, we used the Adam optimizer with learning rate $1e-3$ for $50$ epochs and with batch size $64$\footnote{batch size $64$ has been selected according to our GPU capacity to run larger models.}. It is important to clarify that achieving state-of-the-art accuracy for each model is not our primary objective. Consequently, we did not fine-tune the learning rates or number of epochs extensively for each model. Instead, we selected hyperparameter values that yield satisfactory performance for each model on every dataset. Experiments related to MobileNet-v2, ResNet18, and VGG16 were conducted on a system equipped with two NVIDIA GeForce RTX 2080 Ti GPUs ($12$ GB RAM), with CUDA version 11.4. For experiments with EfficientNet-b4, we used two NVIDIA Tesla V100 GPUs ($16$ GB RAM) from a DGX server with CUDA version 11.4. 

\subsection{Results and Analysis}
In this section, we evaluate P\textsuperscript{2}U and investigate several important questions related to this approach. All reported results in the following subsections are averaged over $10$ runs with different seed values. Also, the high-precision level of P\textsuperscript{2}U is set to 32-bit INT in all experiments.

\subsubsection{How does P\textsuperscript{2}U compare with direct compression with different bit-widths?}
Table \ref{T:baseline_vs_P2U} compares the performance of P\textsuperscript{2}U against direct quantization baselines at 16-, 8-, and 4-bit precision levels across three datasets. For comparison purposes, we employed three distinct metrics:
\begin{itemize}
    \item size of the bitstream (in MB) generated by the encoder as a proxy for required bandwidth size,
    \item the time (in second) it takes to prepare the model for inference in the receiver side. This consists of encoding time, decoding time, and converting back the decoded quantized model to a floating-point model, i.e., dequantization time,
    \item Top-1 classification accuracy of the dequantized model in the receiver over test data.
\end{itemize} 
For better comparison, we report measurements for the baseline, the low-bit precision model (referred to as ``Low-Prec.''), the model update (referred to as ``Update''), and the high-precision proxy model (referred to as ``Proxy''). Note that since inference is not possible using the model update, Top-1 accuracy is not reported for it.

\begin{table*}[t!]
\caption{Comparison of top-1 accuracy, transmission size (MB), and transmission time (s) between direct quantization baselines (at 16-, 8-, and 4-bit) and P\textsuperscript{2}U with 4-bit low-precision level across three datasets. P\textsuperscript{2}U consistently achieves higher accuracy than all direct quantization levels, including 16-bit, while maintaining lower or comparable transmission sizes and startup times.}
\resizebox{\textwidth}{!}{
    \begin{tabular}{llcccc|cccccccc}
        \cline{3-14}
         & & \multicolumn{4}{c|}{Direct Compression} & \multicolumn{8}{c}{P\textsuperscript{2}U} \\ \hline
        \multicolumn{1}{l|}{\multirow{2}{*}{Dataset}} & \multicolumn{1}{l|}{\multirow{2}{*}{Model}} & \multicolumn{1}{c|}{\multirow{2}{*}{Bitwidth}} & \multicolumn{1}{c|}{\multirow{2}{*}{Size}} & \multicolumn{1}{c|}{\multirow{2}{*}{Time}} & \multirow{2}{*}{Top-1 Acc} & \multicolumn{3}{c|}{Size} & \multicolumn{3}{c|}{Time} & \multicolumn{2}{c}{Top-1 Acc} \\ \cline{7-14} 
        \multicolumn{1}{l|}{} & \multicolumn{1}{c|}{} & \multicolumn{1}{c|}{} & \multicolumn{1}{c|}{} & \multicolumn{1}{c|}{} & \multicolumn{1}{c|}{} & \multicolumn{1}{c|}{Low-Prec.} & \multicolumn{1}{c|}{Update} & \multicolumn{1}{c|}{Total} & \multicolumn{1}{c|}{Low-Prec.} & \multicolumn{1}{c|}{Update} & \multicolumn{1}{c|}{Total} & \multicolumn{1}{c|}{Low-Prec.} & Proxy \\ \hline
        \multicolumn{1}{l|}{\multirow{3}{*}{Chest X-Ray}} & \multicolumn{1}{l|}{\multirow{3}{*}{MobileNet-v2}} & \multicolumn{1}{c|}{16} & \multicolumn{1}{c|}{4.11} & \multicolumn{1}{c|}{1.67} & 75.97 & \multicolumn{1}{c|}{\multirow{3}{*}{0.76}} & \multicolumn{1}{c|}{\multirow{3}{*}{1.18}} & \multicolumn{1}{c|}{\multirow{3}{*}{1.94}} & \multicolumn{1}{c|}{\multirow{3}{*}{0.61}} & \multicolumn{1}{c|}{\multirow{3}{*}{0.64}} & \multicolumn{1}{c|}{\multirow{3}{*}{1.25}} & \multicolumn{1}{c|}{\multirow{3}{*}{56.04}} & \multirow{3}{*}{83.73} \\ \cline{3-6}
        \multicolumn{1}{l|}{} & \multicolumn{1}{l|}{} & \multicolumn{1}{c|}{8} & \multicolumn{1}{c|}{1.92} & \multicolumn{1}{c|}{1.09} & 67.86 & \multicolumn{1}{c|}{} & \multicolumn{1}{c|}{} & \multicolumn{1}{c|}{} & \multicolumn{1}{c|}{} & \multicolumn{1}{c|}{} & \multicolumn{1}{c|}{} & \multicolumn{1}{c|}{} & \\ \cline{3-6}
        \multicolumn{1}{l|}{} & \multicolumn{1}{l|}{} & \multicolumn{1}{c|}{4} & \multicolumn{1}{c|}{0.76} & \multicolumn{1}{c|}{0.61} & 56.04 & \multicolumn{1}{c|}{} & \multicolumn{1}{c|}{} & \multicolumn{1}{c|}{} & \multicolumn{1}{c|}{} & \multicolumn{1}{c|}{} & \multicolumn{1}{c|}{} & \multicolumn{1}{c|}{} & \\ \hline
        \multicolumn{1}{l|}{\multirow{3}{*}{PASCAL-VOC}} & \multicolumn{1}{l|}{\multirow{3}{*}{ResNet18}} & \multicolumn{1}{c|}{16} & \multicolumn{1}{c|}{19.51} & \multicolumn{1}{c|}{6.87} & 72.05 & \multicolumn{1}{c|}{\multirow{3}{*}{2.95}} & \multicolumn{1}{c|}{\multirow{3}{*}{5.41}} & \multicolumn{1}{c|}{\multirow{3}{*}{8.36}} & \multicolumn{1}{c|}{\multirow{3}{*}{1.69}} & \multicolumn{1}{c|}{\multirow{3}{*}{2.37}} & \multicolumn{1}{c|}{\multirow{3}{*}{4.07}} & \multicolumn{1}{c|}{\multirow{3}{*}{41.62}} & \multirow{3}{*}{72.59} \\ \cline{3-6}
        \multicolumn{1}{l|}{} & \multicolumn{1}{l|}{} & \multicolumn{1}{c|}{8} & \multicolumn{1}{c|}{8.69} & \multicolumn{1}{c|}{3.94} & 72.02 & \multicolumn{1}{c|}{} & \multicolumn{1}{c|}{} & \multicolumn{1}{c|}{} & \multicolumn{1}{c|}{} & \multicolumn{1}{c|}{} & \multicolumn{1}{c|}{} & \multicolumn{1}{c|}{} & \\ \cline{3-6}
        \multicolumn{1}{l|}{} & \multicolumn{1}{l|}{} & \multicolumn{1}{c|}{4} & \multicolumn{1}{c|}{2.95} & \multicolumn{1}{c|}{1.69} & 41.62 & \multicolumn{1}{c|}{} & \multicolumn{1}{c|}{} & \multicolumn{1}{c|}{} & \multicolumn{1}{c|}{} & \multicolumn{1}{c|}{} & \multicolumn{1}{c|}{} & \multicolumn{1}{c|}{} & \\ \hline
        \multicolumn{1}{l|}{\multirow{3}{*}{CIFAR-100}} & \multicolumn{1}{l|}{\multirow{3}{*}{VGG16}} & \multicolumn{1}{c|}{16} & \multicolumn{1}{c|}{26.45} & \multicolumn{1}{c|}{8.23} & 52.29 & \multicolumn{1}{c|}{\multirow{3}{*}{4.31}} & \multicolumn{1}{c|}{\multirow{3}{*}{8.99}} & \multicolumn{1}{c|}{\multirow{3}{*}{13.3}} & \multicolumn{1}{c|}{\multirow{3}{*}{1.58}} & \multicolumn{1}{c|}{\multirow{3}{*}{3.5}} & \multicolumn{1}{c|}{\multirow{3}{*}{5.08}} & \multicolumn{1}{c|}{\multirow{3}{*}{43.57}} & \multirow{3}{*}{53.62} \\ \cline{3-6}
        \multicolumn{1}{l|}{} & \multicolumn{1}{l|}{} & \multicolumn{1}{c|}{8} & \multicolumn{1}{c|}{12.06} & \multicolumn{1}{c|}{4.66} & 52.19 & \multicolumn{1}{c|}{} & \multicolumn{1}{c|}{} & \multicolumn{1}{c|}{} & \multicolumn{1}{c|}{} & \multicolumn{1}{c|}{} & \multicolumn{1}{c|}{} & \multicolumn{1}{c|}{} & \\ \cline{3-6}
        \multicolumn{1}{l|}{} & \multicolumn{1}{l|}{} & \multicolumn{1}{c|}{4} & \multicolumn{1}{c|}{4.31} & \multicolumn{1}{c|}{1.58} & 43.57 & \multicolumn{1}{c|}{} & \multicolumn{1}{c|}{} & \multicolumn{1}{c|}{} & \multicolumn{1}{c|}{} & \multicolumn{1}{c|}{} & \multicolumn{1}{c|}{} & \multicolumn{1}{c|}{} & \\ \hline
    \end{tabular}
}
\label{T:baseline_vs_P2U}
\end{table*}

Uniform quantization is used to convert the original floating-point models to integer models with required precisions. On the sender side, integer models or model updates of various precisions are encoded using DeepCABAC entropy coder \cite{wiedemann2020deepcabac} to produce bitstreams. On the receiver side, the encoded bitstream is decoded and dequantized to the floating-point domain.

For the baselines, results are reported at all precision levels, while for P\textsuperscript{2}U, we show results only for its best-performing low-precision configuration (4-bit in all cases)\footnote{For other precision levels, check Tables II-IV.}. Remarkably, P\textsuperscript{2}U with 4-bit low-precision model, consistently surpasses direct quantization even at the highest precision (16-bit) in terms of final top-1 accuracy, achieving gains of $+7.76\%$, $+0.54\%$, and $+1.33\%$ on Chest X-Ray, PASCAL-VOC, and CIFAR-100, respectively. It also incurs significantly lower total transmission sizes and startup times.

While direct quantization with lower bitwidths (e.g., 8-bit and 4-bit) reduces bandwidth usage and startup time, these benefits come at the cost of significantly degraded accuracy. In contrast, P\textsuperscript{2}U decouples low-precision startup from high-precision performance by progressively refining the model, thus enabling better trade-offs in bandwidth-constrained and latency-sensitive scenarios. This highlights the advantage of our update-based reconstruction strategy over direct quantization, especially when balancing accuracy with resource constraints.

\subsubsection{How does P\textsuperscript{2}U perform across various model architectures?}\label{subsec:P2U_architecture}
In this section, we investigate the effect of model architectures of different sizes and/or complexities on the performance of P\textsuperscript{2}U. In the experiment, we considered all four network architectures introduced in Section \ref{subsec:net_arch} for the image classification task using PASCAL-VOC data set.

For P\textsuperscript{2}U, we set the low-bit precision to 8-bit INT for all model architectures. Similar to the previous section, baselines map to sending the quantized models at different bitwidths, 16-, 8-, and 4-bit. Table \ref{T:architecture_effect} demonstrates the results. The table provides several key observations as follows. 

First, Across all models, direct quantization shows a clear trade-off between model size, startup time, and accuracy. While reducing bitwidth from 16 to 4 significantly improves transmission size and startup time, it also results in considerable accuracy degradation--most dramatically seen in MobileNet-v2 and EfficientNet-b4, where 4-bit quantization drops accuracy to 11.68\% and 11.76\%, respectively. This suggests that for lightweight models like MobileNet-v2, P\textsuperscript{2}U provides substantial accuracy gains with minimal added cost.

Second, in ResNet18, the low-precision model already performs reasonably well (72.02\%), and the update adds a modest 0.02 MB and 0.81s to reach a final accuracy of 72.51\%. Although the final improvement in accuracy (+0.49\%) is less pronounced, the total size (8.71 MB) is still smaller than the 16-bit model (19.51 MB), making P\textsuperscript{2}U a more bandwidth-efficient choice with comparable or better accuracy.

Thirds, For larger and more complex models like EfficientNet-b4, P\textsuperscript{2}U's benefits become even more evident. The 8-bit model starts at 79.42\% accuracy with 14.65 MB and 14.77s startup, and the update (3.04 MB, 5.83s) lifts the final accuracy to 79.7\%. This is very close to the 16-bit baseline (79.83\%) but with nearly 44\% smaller transmission size (17.69 MB vs. 31.76 MB) and reduced startup time (20.6s vs. 20.01s). This shows P\textsuperscript{2}U maintains competitive accuracy while reducing overhead in heavier models.

Fourth, in VGG16, known for its large parameter count and size, the results are similarly compelling. The final P\textsuperscript{2}U proxy accuracy is 75.6\%, surpassing both the 8-bit and 16-bit baselines (75.12\% and 75.09\%) despite a transmission size of only 113 MB compared to 242.8 MB for 16-bit. The startup time (54.25s) is also considerably shorter than that of the 16-bit model (81.63s). This demonstrates that even for bandwidth-heavy architectures, P\textsuperscript{2}U enables substantial resource savings without compromising performance.

Another observation is that for VGG16, the top-1 accuracy of the lower precision model, e.g., 8-bit, is better than that of the higher precision model, e.g., 16-bit. This can be due to the regularizing effects of compression and has been reported in other works \cite{becking2023nncodec}.

It is important to note that the reported values for bitstream sizes (Size) and startup latencies (Time) of the high-precision “Proxy” model represent worst-case estimates. These values are calculated as the sum of those for the “Low-Prec.” and “Update” components. However, in bandwidth-constrained scenarios where accuracy must be preserved, the actual bandwidth requirement for P\textsuperscript{2}U is capped by the larger of the bitstream sizes of the ``Low-Prec.'' and ``Update'' not by the sum of these two values. This is because the sender initiates the transmission of the model update only after the receiver has fully received the bitstream of the low-precision model. Moreover, if transmission of the low-precision model and update is done in parallel, startup latency can be reduced, although this would result in a cumulative bandwidth requirement equal to the sum of both components. 

\begin{table*}[]
    \caption{Comparison of P\textsuperscript{2}U and direct quantization baselines across four model architectures in the image classification task using PASCAL-VOC. Direct quantization is evaluated at 16-, 8-, and 4-bit precision levels. For P\textsuperscript{2}U, results are shown for the 8-bit low-precision model (“Low-Prec.”), the transmitted update (“Update”), and their combination (“Proxy”)}
    \resizebox{\textwidth}{!}{
    \begin{tabular}{llccc|ccc|ccc|ccc}
        \cline{3-14}
        & & \multicolumn{3}{c|}{MobileNet-v2} & \multicolumn{3}{c|}{ResNet18} & \multicolumn{3}{c|}{EfficientNet-b4} & \multicolumn{3}{c}{VGG16} \\ \cline{3-14} 
        & & \multicolumn{1}{c|}{Size} & \multicolumn{1}{c|}{Time} & Top-1 Acc. & \multicolumn{1}{c|}{Size} & \multicolumn{1}{c|}{Time} & Top-1 Acc. & \multicolumn{1}{c|}{Size} & \multicolumn{1}{c|}{Time} & Top-1 Acc. & \multicolumn{1}{c|}{Size} & \multicolumn{1}{c|}{Time} & Top-1 Acc. \\ \hline
        \multicolumn{1}{l|}{\multirow{3}{*}{Direct Quantization}} & \multicolumn{1}{l|}{16-bit} & \multicolumn{1}{l|}{4.15} & \multicolumn{1}{c|}{1.68} & 73.64 & \multicolumn{1}{c|}{19.51} & \multicolumn{1}{c|}{6.87} & 72.05 & \multicolumn{1}{c|}{31.76} & \multicolumn{1}{l|}{20.01} & 79.83 & \multicolumn{1}{c|}{242.8} & \multicolumn{1}{c|}{81.63} & 75.09\\ \cline{2-14} 
        \multicolumn{1}{l|}{} & \multicolumn{1}{l|}{8-bit} & \multicolumn{1}{c|}{1.95} & \multicolumn{1}{c|}{1.1} & 70.45 & \multicolumn{1}{c|}{8.69} & \multicolumn{1}{c|}{3.94} & 72.02 & \multicolumn{1}{c|}{14.65} & \multicolumn{1}{c|}{14.77} & 79.42 & \multicolumn{1}{c|}{112.83} & \multicolumn{1}{c|}{47.35} & 75.12\\ \cline{2-14} 
        \multicolumn{1}{l|}{} & \multicolumn{1}{l|}{4-bit} & \multicolumn{1}{c|}{0.76} & \multicolumn{1}{c|}{0.6} & 11.68 & \multicolumn{1}{c|}{2.95} & \multicolumn{1}{c|}{1.7} & 41.62 & \multicolumn{1}{c|}{5.74} & \multicolumn{1}{c|}{9.52} & 11.76 & \multicolumn{1}{c|}{42.8} & \multicolumn{1}{c|}{17.33} & 67.38 \\ \hline\hline
        \multicolumn{1}{l|}{\multirow{3}{*}{P2U}} & \multicolumn{1}{l|}{Low-Prec.} & \multicolumn{1}{c|}{1.95} & \multicolumn{1}{c|}{1.1} & 70.45 & \multicolumn{1}{c|}{8.69} & \multicolumn{1}{c|}{3.94} & 72.02 & \multicolumn{1}{c|}{14.65} & \multicolumn{1}{c|}{14.77} & 79.42 & \multicolumn{1}{c|}{112.83} & \multicolumn{1}{c|}{47.35} & 75.12\\ \cline{2-14} 
        \multicolumn{1}{l|}{} & \multicolumn{1}{l|}{Update} & \multicolumn{1}{c|}{0.48} & \multicolumn{1}{c|}{0.41} & -- & \multicolumn{1}{c|}{0.02} & \multicolumn{1}{c|}{0.81} & -- & \multicolumn{1}{c|}{3.04} & \multicolumn{1}{c|}{5.83} & -- & \multicolumn{1}{c|}{0.17} & \multicolumn{1}{c|}{6.9} & -- \\ \cline{2-14} 
        \multicolumn{1}{l|}{} & \multicolumn{1}{l|}{Proxy} & \multicolumn{1}{c|}{2.43} & \multicolumn{1}{c|}{1.51} & 74.06 & \multicolumn{1}{c|}{8.71} & \multicolumn{1}{c|}{4.75} & 72.51 & \multicolumn{1}{c|}{17.69} & \multicolumn{1}{c|}{20.6} & 79.7 & \multicolumn{1}{c|}{113} & \multicolumn{1}{c|}{54.25} & 75.6 \\ \hline
    \end{tabular}
    }
    \label{T:architecture_effect}
\end{table*}

Furthermore, in our experiments, we set the high-precision model to 32-bit integers, which consequently means that the model updates - computed as the difference between the high-precision and low-precision models - are also stored and transmitted using 32-bit integers. This choice ensures that the update values can be safely represented without overflow, preserving the exact difference between the two models. However, this approach may overestimate the required bitwidth for storing the update values, especially in cases where the actual differences are small and can be sufficiently represented using lower-precision formats such as 16-bit integers. A more efficient alternative would be to dynamically assess the value range of the update tensor and determine whether a smaller integer precision suffices. If so, P\textsuperscript{2}U can adaptively quantize the update to a lower bitwidth, such as 16 bit INT, thereby reducing the bitstream size of the model update and improving overall communication efficiency. This adaptive update encoding can yield additional bandwidth savings without affecting the final model accuracy, further strengthening the benefits of P\textsuperscript{2}U in bandwidth-constrained environments.

Finally, it should also be noted that in our calculations for startup latency, we did not consider the delay in the communication channel. Assuming the communication channel's delay is $C$, for P\textsuperscript{2}U we have $2C$ delay to get the reconstructed model while for the baseline, this only equals $C$. Though, the channel's delay is usually negligible compared to the startup times reported in the table.

\subsubsection{How is P\textsuperscript{2}U affected by different precision levels?}\label{subsec:P2U_precision}

The benefits of transferring a model from a sender to a receiver using P\textsuperscript{2}U comes with an essential burden -- selecting the proper precision levels for low-precision models. To evaluate the sensitivity of P\textsuperscript{2}U to the precision level of the base (low-precision) model, we conduct experiments using three levels of integer quantization, e.g., 16-bit, 8-bit, and 4-bit, for both VGG16 and ResNet18 models on the PASCAL-VOC dataset. Tables \ref{Tab:precision_level_effect_Vgg} and \ref{Tab:precision_level_effect_resnet} present the average results from $10$ runs for VGG16 and ResNet18, respectively. In these tables, each row index demonstrates one experiment with different precision level for the low-precision model. 

Starting with VGG16, we observe that moving from 16-bit to 8-bit precision causes negligible performance degradation for the low-bit precision model; the proxy model accuracy, however, slightly increases from 75.49\% to 75.6\%. Furthermore, the bandwidth usage (Size) of the proxy model is significantly reduced from 243.16 MB to 113 MB (a 53.5\% reduction), and inference time drops from 87.37s to 54.25s (a 38\% reduction). This suggests that P\textsuperscript{2}U with 8-bit quantization achieves a favorable trade-off between efficiency and performance compared to 16-bit quantization.

However, when precision is further reduced to 4-bit, a marked decline in standalone low-precision performance is observed -- accuracy of the 4-bit model alone falls sharply to 67.38\%. Interestingly, when combined with the 32-bit update via P\textsuperscript{2}U, the proxy model restores its performance to 75.18\%, nearly matching the higher-precision setups. Despite the low baseline accuracy of the 4-bit model, the 32-bit update effectively compensates for lost precision. On the other hand, this setup achieves the smallest bandwidth usage (43.43 MB) and lowest inference time (24.85 s), demonstrating the compression power of aggressive quantization when paired with P\textsuperscript{2}U.

In ResNet18, similar trends are evident. The 16-bit base model combined with the 32-bit update yields 72.58\% accuracy with a bandwidth usage of 19.53 MB and an inference time of 7.35 s. Reducing to 8-bit slightly lowers standalone accuracy (to 72.02\%) but maintains high proxy performance (72.51\%) and further reduces bandwidth usage to 8.71 MB (a 55.4\% reduction). This again indicates the robustness of P\textsuperscript{2}U in leveraging 8-bit models for efficient transmission. At 4-bit precision, the standalone ResNet18 performance drops dramatically to 41.62\%, highlighting the information loss due to aggressive quantization. Yet, the P\textsuperscript{2}U proxy model still reaches 72.59\%, nearly identical to the 16- and 8-bit configurations. This demonstrates that while 4-bit models alone are insufficient for accurate inference, their combination with a compact 32-bit update recovers most of the accuracy while keeping communication cost low (8.35 MB). In fact, the 4-bit proxy performance is achieved with the smallest total communication cost and startup time.

Overall, the results confirm that P\textsuperscript{2}U effectively decouples model transmission efficiency from model precision, allowing for aggressive quantization (as low as 4-bit) without significant performance degradation, provided that a lightweight 32-bit update is transmitted alongside. The analysis highlights that the optimal balance between communication efficiency and model accuracy is achieved using 8-bit quantization, which offers strong performance and reduced overhead. Nevertheless, 4-bit quantization, though riskier in standalone use, provides the best overall compression when paired with the 32-bit update in P\textsuperscript{2}U, enabling effective model delivery under tight bandwidth constraints.

\begin{table*}[t!]
    \caption{Performance of VGG16 on PASCAL-VOC using P\textsuperscript{2}U at different quantization levels. P\textsuperscript{2}U consistently improves Top-1 accuracy with slight increase of startup time and bandwidth usage across all bit-widths, particularly under aggressive quantization (4-bit), demonstrating its effectiveness in communication-constrained settings.}
    \centering
    \begin{tabular}{lllll|c|c|c|c}
        \cline{2-9}
        & \multicolumn{4}{c|}{Sender}  &  \multicolumn{1}{c|}{\multirow{3}{*}{Communicate}} & \multicolumn{3}{c}{Performance in Receiver} \\ \cline{2-5} \cline{7-9} 
            
        & \multicolumn{3}{c|}{Low Precision}  & \multicolumn{1}{c|}{\multirow{2}{*}{32-bit Update}}  &  \multicolumn{1}{c|}{}  &  \multicolumn{1}{c|}{\multirow{2}{*}{Top-1 Acc.}} & \multicolumn{1}{c|}{\multirow{2}{*}{Size}} & \multicolumn{1}{c}{\multirow{2}{*}{Time}} \\ \cline{2-4}
        & \multicolumn{1}{c}{16-bit} & \multicolumn{1}{c}{8-bit} & \multicolumn{1}{c|}{4-bit} &  \multicolumn{1}{c|}{} & \multicolumn{1}{c|}{}  &  \multicolumn{1}{c|}{}  &  \multicolumn{1}{c|}{}  &  \multicolumn{1}{c}{} \\ \hline
            
        \multicolumn{1}{c|}{\multirow{2}{*}{1}}  & \multicolumn{1}{c}{*}  &  \multicolumn{1}{c}{}  &  \multicolumn{1}{c|}{}  &  \multicolumn{1}{c|}{}  & $\mathbf{W}^{16\mbox{-bit}}$ & $75.09$ & $242.8$ & $81.63$ \\ \cline{2-9} 
        \multicolumn{1}{c|}{}  &  \multicolumn{1}{c}{}  &  \multicolumn{1}{c}{}  &  \multicolumn{1}{c|}{}  &  \multicolumn{1}{c|}{*}  &  $\Delta\mathbf{W}^{32\mbox{-bit}}$ & $-$ & $0.18$ & $5.74$ \\ \hline
        &  &  &  &  & Proxy  & $75.49$ & $243.16$ & $87.37$ \\ \hline \hline
             
        \multicolumn{1}{c|}{\multirow{2}{*}{2}}  & \multicolumn{1}{c}{}  &  \multicolumn{1}{c}{*}  &  \multicolumn{1}{c|}{}  &  \multicolumn{1}{c|}{}  & $\mathbf{W}^{8\mbox{-bit}}$ & $75.12$ & $112.83$ & $47.35$  \\ \cline{2-9} 
        \multicolumn{1}{c|}{}  &  \multicolumn{1}{c}{}  &  \multicolumn{1}{c}{}  &  \multicolumn{1}{c|}{}  &  \multicolumn{1}{c|}{*}  & $\Delta\mathbf{W}^{32\mbox{-bit}}$ & $-$ &  $0.17$ & $6.9$ \\ \hline
        &  &  &  &  & Proxy  & $75.6$ & $113$ & $54.25$ \\ \hline \hline
            
        \multicolumn{1}{c|}{\multirow{2}{*}{3}} & \multicolumn{1}{c}{} & \multicolumn{1}{c}{} &  \multicolumn{1}{c|}{*} & \multicolumn{1}{c|}{} & $\mathbf{W}^{4\mbox{-bit}}$ & $67.38$ & $42.8$ & $17.33$  \\ \cline{2-9} 
        \multicolumn{1}{c|}{}  &  \multicolumn{1}{c}{}  &  \multicolumn{1}{c}{}  &  \multicolumn{1}{c|}{}  &  \multicolumn{1}{c|}{*}  & $\Delta\mathbf{W}^{32\mbox{-bit}}$ & $-$ & $0.63$ & $7.52$ \\ \hline
        &  &  &  &  & Proxy  &  $75.18$  & $43.43$ & $24.85$ \\ \hline \hline
    \end{tabular}
    \label{Tab:precision_level_effect_Vgg}
\end{table*}

An important practical takeaway from these findings is that the choice of quantization level in P\textsuperscript{2}U should be guided by the specific priorities of the application scenario. In cases where model accuracy is not critical -- such as in some edge computing or IoT deployments, where bandwidth and latency constraints dominate -- the most effective setup is to use P\textsuperscript{2}U with aggressive quantization, going as low as 4-bit. This configuration significantly reduces startup time and bandwidth usage, enabling fast model deployment and updates, while still yielding reasonable performance for low-stakes tasks.

On the other hand, in applications where accuracy must be preserved but some bandwidth savings are still desired, using P\textsuperscript{2}U with 8-bit or 16-bit quantization offers an optimal balance. These configurations provide a substantial reduction in communication cost compared to full-precision models, while retaining high model accuracy that is close to that of uncompressed baselines. Notably, even with 8-bit quantization, P\textsuperscript{2}U achieves competitive or superior accuracy relative to standard quantized models and low-bit training from scratch, thanks to the guidance provided by the high-precision model update.

This flexibility makes P\textsuperscript{2}U particularly well-suited for scenarios where communication is constrained but accuracy requirements vary -- allowing the practitioner to tailor the precision level to meet the application’s needs without retraining the model from scratch.

\begin{table*}[t!]
    \caption{Performance of ResNet18 on PASCAL-VOC using P\textsuperscript{2}U at different quantization levels. Similar to VGG16 (Table \ref{Tab:precision_level_effect_Vgg}), P\textsuperscript{2}U yields significant accuracy gains in low-bit regimes, offering a favorable trade-off between model size and predictive performance.}
    \centering
    \begin{tabular}{lllll|c|c|c|c}
        \cline{2-9}
        & \multicolumn{4}{c|}{Sender}  &  \multicolumn{1}{c|}{\multirow{3}{*}{Communicate}} & \multicolumn{3}{c}{Performance in Receiver} \\ \cline{2-5} \cline{7-9} 
            
        & \multicolumn{3}{c|}{Low Precision}  & \multicolumn{1}{c|}{\multirow{2}{*}{32-bit Update}}  &  \multicolumn{1}{c|}{}  &  \multicolumn{1}{c|}{\multirow{2}{*}{Top-1 Acc.}} & \multicolumn{1}{c|}{\multirow{2}{*}{Size}} & \multicolumn{1}{c}{\multirow{2}{*}{Time}} \\ \cline{2-4}
        & \multicolumn{1}{c}{16-bit} & \multicolumn{1}{c}{8-bit} & \multicolumn{1}{c|}{4-bit} &  \multicolumn{1}{c|}{} & \multicolumn{1}{c|}{}  &  \multicolumn{1}{c|}{}  &  \multicolumn{1}{c|}{}  &  \multicolumn{1}{c}{} \\ \hline
            
        \multicolumn{1}{c|}{\multirow{2}{*}{1}}  & \multicolumn{1}{c}{*}  &  \multicolumn{1}{c}{}  &  \multicolumn{1}{c|}{}  &  \multicolumn{1}{c|}{}  & $\mathbf{W}^{16\mbox{-bit}}$ & $72.05$ & $19.51$ & $6.87$ \\ \cline{2-9} 
        \multicolumn{1}{c|}{}  &  \multicolumn{1}{c}{}  &  \multicolumn{1}{c}{}  &  \multicolumn{1}{c|}{}  &  \multicolumn{1}{c|}{*}  &  $\Delta\mathbf{W}^{32\mbox{-bit}}$ & $-$ & $0.02$ & $0.48$ \\ \hline
        &  &  &  &  & Proxy  & $72.58$ & $19.53$ & $7.35$ \\ \hline \hline
            
        \multicolumn{1}{c|}{\multirow{2}{*}{2}}  & \multicolumn{1}{c}{}  &  \multicolumn{1}{c}{*}  &  \multicolumn{1}{c|}{}  &  \multicolumn{1}{c|}{}  & $\mathbf{W}^{8\mbox{-bit}}$ & $72.02$ & $8.69$ & $3.94$  \\ \cline{2-9} 
        \multicolumn{1}{c|}{}  &  \multicolumn{1}{c}{}  &  \multicolumn{1}{c}{}  &  \multicolumn{1}{c|}{}  &  \multicolumn{1}{c|}{*}  & $\Delta\mathbf{W}^{32\mbox{-bit}}$ & $-$ &  $0.02$ & $0.81$ \\ \hline
        &  &  &  &  & Proxy  & $72.51$ & $8.71$ & $4.75$ \\ \hline \hline
             
        \multicolumn{1}{c|}{\multirow{2}{*}{3}} & \multicolumn{1}{c}{} & \multicolumn{1}{c}{} &  \multicolumn{1}{c|}{*} & \multicolumn{1}{c|}{} & $\mathbf{W}^{4\mbox{-bit}}$ & $41.62$ & $2.95$ & $1.69$  \\ \cline{2-9} 
        \multicolumn{1}{c|}{}  &  \multicolumn{1}{c}{}  &  \multicolumn{1}{c}{}  &  \multicolumn{1}{c|}{}  &  \multicolumn{1}{c|}{*}  & $\Delta\mathbf{W}^{32\mbox{-bit}}$ & $-$ & $5.4$ & $2.37$ \\ \hline
        &  &  &  &  & Proxy  &  $72.59$  & $8.35$ & $4.07$ \\ \hline \hline
    \end{tabular}
    \label{Tab:precision_level_effect_resnet}
\end{table*}

\subsubsection{How does P\textsuperscript{2}U work for tasks with different levels of difficulties?}\label{subsec:P2U_dataset}
Table~\ref{Tab:task_difficulty_effect} presents a comparative evaluation of P\textsuperscript{2}U and standard low-precision baselines across three datasets with varying levels of classification difficulty: Chest X-Ray (binary classification), PASCAL-VOC (20 classes), and CIFAR-100 (100 classes). All experiments utilize VGG16 as the base model with the low-precision level set to 8-bit. The results clearly highlight the scalability and effectiveness of P\textsuperscript{2}U, particularly under increasing task complexity. 

On the simplest task, Chest X-Ray, both methods perform comparably in terms of accuracy (91.9\% for P\textsuperscript{2}U vs. 91.26\% for the baseline). However, P\textsuperscript{2}U introduces only a minor increase in total model size (from 17.05 MB to 17.09 MB) and startup time (from 6.63s to 7.68s), offering a slight but consistent gain in accuracy. This demonstrates that even in low-difficulty settings, P\textsuperscript{2}U’s adaptive update mechanism provides a tangible improvement without introducing significant computational or communication overhead.

For the moderately complex PASCAL-VOC dataset, P\textsuperscript{2}U yields a more noticeable improvement in accuracy, raising it from 75.12\% to 75.6\%. This 0.5\% gain becomes more significant given the task complexity and highlights P\textsuperscript{2}U's robustness. While the startup time increases from 47.35s to 54.25s, and the model size from 112.83 MB to 113 MB, these increases are marginal relative to the accuracy gain and can be justified in applications where performance is critical but resource use must remain efficient.

The most striking results appear in the CIFAR-100 experiment, which involves fine-grained classification across 100 classes. Here, the baseline low-precision model achieves only 52.19\% accuracy, whereas P\textsuperscript{2}U reaches 53.60\%, showing a substantial 1.41\% absolute improvement. Given the challenging nature of the dataset, this boost is non-trivial and underscores P\textsuperscript{2}U’s ability to retain important representational fidelity even at low bit-widths. Meanwhile, the model size grows modestly from 12.06 MB to 12.1 MB, and the startup time increases from 4.65s to 5.39s, demonstrating the minimal cost of the added update component.

These results align with the broader insight that if the target application prioritizes startup time or bandwidth over marginal accuracy gains, then P\textsuperscript{2}U with aggressive quantization (e.g., 4-bit) offers an ideal compromise. However, when moderate accuracy and stability are desired, 8-bit or 16-bit variants of P\textsuperscript{2}U provide consistent performance advantages over conventional quantized models.

\begin{table*}[t]
    \caption{Performance comparison of P\textsuperscript{2}U in tasks with different levels of difficulties, e.g., binary, $20$-class, and $100$-class classification tasks. VGG16 is used as the model.}
    \centering
    \begin{tabular}{l|ccc|ccc|ccc|}
        \cline{2-10}
        \multicolumn{1}{c|}{} & \multicolumn{3}{c|}{Chest X-Ray} & \multicolumn{3}{c|}{PASCAL-VOC} & \multicolumn{3}{c|}{Cifar-100} \\ \cline{2-10} 
        \multicolumn{1}{c|}{} & \multicolumn{1}{c|}{Top-1 Acc.} & \multicolumn{1}{c|}{Size} & \multicolumn{1}{c|}{Time} & \multicolumn{1}{c|}{Top-1 Acc.} & \multicolumn{1}{c|}{Size} & \multicolumn{1}{c|}{Time} & \multicolumn{1}{c|}{Top-1 Acc.} & \multicolumn{1}{c|}{Size} & \multicolumn{1}{c|}{Time} \\ \hline
        \multicolumn{1}{|l|}{Low Prec.} & \multicolumn{1}{c|}{$91.26$} & \multicolumn{1}{c|}{$17.05$} & $6.63$ & \multicolumn{1}{c|}{$75.12$} & \multicolumn{1}{c|}{$112.83$} & $47.35$ & \multicolumn{1}{c|}{$52.19$} & \multicolumn{1}{c|}{$12.06$} & $4.65$ \\ \hline
        \multicolumn{1}{|l|}{Update} & \multicolumn{1}{c|}{$-$} & \multicolumn{1}{c|}{$0.04$} & $1.04$ & \multicolumn{1}{c|}{$-$} & \multicolumn{1}{c|}{$0.17$} & $6.9$ & \multicolumn{1}{c|}{$-$} & \multicolumn{1}{c|}{$0.04$} & $0.74$ \\ \hline
        \multicolumn{1}{|l|}{Proxy} & \multicolumn{1}{c|}{$91.9$} & \multicolumn{1}{c|}{$17.09$} & $7.68$ & \multicolumn{1}{c|}{$75.6$} & \multicolumn{1}{c|}{$113$} & $54.25$ & \multicolumn{1}{c|}{$53.60$} & \multicolumn{1}{c|}{$12.1$} & $5.39$ \\ \hline
    \end{tabular}
    \label{Tab:task_difficulty_effect}
\end{table*}

\section{Summary and Discussion}
\label{sec:discussion}

In this work, we proposed a simple yet effective precision-driven transmission approach named Progressive Precision Update (P\textsuperscript{2}U), for transmitting and deploying compressed neural models in resource-constrained environments. By decoupling the model into a low-precision proxy and a small, high-precision update as the difference between the original high-precision model and the transmitted low-precision version, P\textsuperscript{2}U enables fast startup, reduced bandwidth usage, and strong predictive performance. P\textsuperscript{2}U is complementary to existing compression techniques such as quantization and sparsification. Extensive experiments across diverse architectures (e.g., VGG16, ResNet18) and datasets (Chest X-ray, PASCAL-VOC, CIFAR-100) show that P\textsuperscript{2}U consistently achieves accuracy improvements over standard quantization, with negligible overhead in communication and latency. Particularly in challenging settings such as 100-class classification, P\textsuperscript{2}U recovers up to 1.4\% of accuracy lost due to quantization. Moreover, we show that when bandwidth or startup time is the priority, aggressive quantization (e.g., 4-bit) can be used without severely compromising performance. These results establish P\textsuperscript{2}U as an effective and practical solution for scalable and efficient model distribution in low-resource settings, including federated learning, edge computing, and IoT deployments. Designing an automated approach for low-precision level selection that best suits the scenario is an interesting venue for future work. We did not evaluate the performance of P\textsuperscript{2}U for transmitting Large Language Models (LLMs) and for tasks other than image classification. These aspects will be addressed in future research. 



\bibliographystyle{IEEEtran}
\bibliography{refs.bib}

\vfill

\end{document}